# Article

## Prediction of Rapid Early Progression and Survival Risk with Pre-Radiation MRI in WHO Grade 4 Glioma Patients

Walia Farzana [1], Mustafa M Basree [2], Norou Diawara [3], Zeina A. Shboul [1], Sagel Dubey [2], Marie M Lockhart [4], Mohamed Hamza [5], Joshua D. Palmer [6], Khan M. Iftekharuddin [1,*]

[1] Vision Lab, Department of Electrical & Computer Engineering, Old Dominion University, Norfolk, VA, USA
[2] Department of Internal Medicine, OhioHealth Riverside Methodist Hospital, Columbus, OH, USA
[3] Department of Mathematics & Statistics, Old Dominion University, Norfolk, VA, USA
[4] OhioHealth Research Institute, Columbus, OH, USA
[5] Department of Neurology, OhioHealth, Columbus, OH, USA
[6] Department of Radiation Oncology, The James Cancer Hospital and Solove Research Institute, Ohio State University Wexner Medical Center, Columbus, OH, USA
* Correspondence: kiftekha@odu.edu

**Simple Summary:** Rapid Early Progression (REP) has been defined as increased nodular enhancement at the border of the resection cavity, the appearance of new lesions outside the resection cavity, or increased enhancement of the residual disease after surgery and before radiation. Patients with REP have worse survival compared to patients without REP (non-REP), therefore, a reliable method for differentiating REP from non-REP is hypothesized to assist in persolnized tratement planning. A potential approach is to use the radiomics and fractal texture features extracted from brain tumors to characterize morphological and physiological properties. We propose a random sampling-based ensemble classification model. The proposed iterative random sampling of patient data followed by feature selection and classification method with radiomics, multi-resolution fractal, and proteomics features predict REP from non-REP using radiation-planning magnetic resonance imaging (MRI). Our results further show efficacy of pre-radiation image features in the analysis of survival probability and prognostic grouping of patients.

**Abstract:** Recent clinical research describes a subset of glioblastoma patients that exhibit REP prior to start of radiation therapy. Current literature has thus far described this population using clinico-pathologic features. To our knowledge, this study is the first to investigate the potential of conventional radiomics, sophisticated multi-resolution fractal texture features, and different molecular features (MGMT, IDH mutations) as a diagnostic and prognostic tool for prediction of REP from non-REP cases using computational and statistical modeling methods. Radiation-planning T1 post-contrast (T1C) MRI sequences of 70 patients are analyzed. Ensemble method with 5-fold cross validation over 1000 iterations offers AUC of $0.793\pm0.082$ for REP and non-REP classification. In addition, copula-based modeling under dependent censoring (where a subset of the patients may not be followed up until death) identifies significant features (p-value <0.05) for survival probability and prognostic grouping of patient cases. The prediction of survival for the patient's cohort produces precision of $0.881\pm0.056$. The prognostic index (PI) calculated using the fused features suggests that 84.62% of REP cases fall under the bad prognostic group, suggesting potentiality of fused features to predict a higher percentage of REP cases. The experimental result further shows that multi-resolution fractal texture features perform better than conventional radiomics features for REP and survival outcomes.

## 1. Introduction

Brain and other central nervous system (CNS) tumors are associated with highest mortality and morbidity across different malignancies in the United States [1]. Glioblastoma (GB) is one of the most aggressive brain tumors representing nearly half of brain gliomas [2]. For newly diagnosed GB patients, standard of care includes maximal safe surgical resection followed by radiation therapy with concurrent and adjuvant temozolomide with Novo-TTF [3]. Radiation therapy should ideally begin six weeks following surgery [4]. During this time frame GB may regrow significantly due to its highly proliferative nature [5]. Several institutional series have been published evaluating post-operative REP with pre-radiation MRI [6–8]. Specifically, REP is evaluated by comparing early postoperative MRI with radiation-planning MRI [8]. There is an almost 50% prevalence rate for development of REP, even if radiation is initiated earlier than six weeks post-operatively [6–8].

MRI plays a crucial role in evaluation of post-operative and post-treatment effects. Post-operative MRI is acquired within 3 days (preferably within 24 hours) both to assess extent of resection and to minimize the effect of enhancement due to surgery [5]. Radiation planning MRI is obtained 1-3 weeks prior to start of radiation to assist in target delineation. According to the guidelines [9], another MRI (post-radiation) is obtained 2-6 weeks after completion of treatment with radiation therapy +/- TMZ followed by surveillance imaging every 2-4 months. The first post-radiation MRI is usually compared to early post-operative, or more commonly to pre-radiation, MRI to evaluate tumor progression and/or radiation-related changes (collectively termed pseudo progression) [4]. MRI offers detailed characterizations of the morphological, physiological, and metabolic properties of brain tumors, particularly brain gliomas, which are complex and heterogenous malignancies [10]. Quantitative radiomics features [11,12] extracted from MRI (intensity, shape, area, texture, and geometric) followed by statistical and machine learning analyses have shown to be effective [13–15] in brain tumor volume segmentation, and classification of normal/tumor tissues.

Patients with REP has worse survival compared to non-REP [7]. Overall survival (OS) analysis refers to the time-to-death from day of surgery. A patient is censored if they are lost to follow-up prior to observing the time-to-death [16]. Censoring may induce bias into statistical analysis results if censoring processes involve dropout or withdrawal due to tumor progression, treatment toxicity, or start of second-line therapy [16]. Because a patient may die soon after being lost follow-up, total survival and dropout time may be positively associated [16]. Dependent censoring occurs when the relationship between censoring time and survival time cannot be explained by observable factors [16,17]. In statistical analysis, copula-based modeling is a state-of-the-art method to model the dependency between survival and censoring time.

As alluded to above, several retrospective reviews correlated clinical and pathologic features with REP, and found it to be an independent negative prognostic factor [4], [6–8]. It remains unclear if patients with REP have distinct molecular or radiographic features. To the best of our knowledge, there has been no research that utilizes MRI imaging features for prediction of REP using radiation-planning MRI, which may act as a quantitative imaging-based biomarker to stratify REP patients from non-REP.

The first objective of the study is to assess the prediction efficacy of conventional radiomics and sophisticated multi-resolution fractal texture feature extracted from radiation-planning MRI to predict REP. A second objective is to analyze the survival probability of patients using copula modeling of radiation-planning MRI radiomics features under dependent censoring scenario. Third objective is the binary prediction of patients survival (patients expired/dead or not) from selected significant (p-value<0.005) T1C features utilizing copula modeling. To evaluate the significance of the selected features, prognostic index (PI) is calculated as linear combinations of the selected features. Based on the calculated PI, prognostic grouping (good or bad) is obtained. Within the prognostic group the distribution of REP vs. non-REP cases is also developed. The distribution of REP vs non-REP cases within the prognostic groups further signifies the predcitive ability of radiomic features in identifying REP as a bad prognostic or high risk group.

## 2. Materials and Methods

*2.1 Patient Data*

This is an institutional review board (IRB) approved retrospective chart review of a cohort of patients treated at Ohio-Health between 1/1/2015 and 3/1/2021. Relevant clinical and radiographic data was abstracted from electronic health record system. Patients with biopsy-confirmed diagnosis of World Health Organization (WHO) grade 3 or 4 anaplastic astrocytoma or grade 4 glioblastoma were included in the study, with at least three MRI (pre-operative, early post-operative, radiation planning). Imaging studies were reviewed by board-certified neuroradiologists. All patients underwent surgery (biopsy, subtotal, or gross total resection) followed by radiation therapy with or without adjuvant temozolomide.

A total of 95 patient cases were included in this study. Among these, twenty-five cases did not have (T1C) sequence in at least one of the MRI studies. Seventy patients with complete radiographic data were included in the analysis, thirteen of which were clinically identified as REP with remaining fifty-seven as non-REP. REP was classified as such in line with previous literature [18,19]. A detailed summary of patient's cases is presented in Table 1.

**Table 1**. Summary of Data between the patient group

|  | Total (n=70) | REP (n=13) | Non-REP(n=57) |
|---|---|---|---|
| **Survival Days from Surgery** | | | |
| Present (Patient dead/expired) | 45 | 8 | 37 |
| Lost Follow-up (Censored)* | 22 | 5 | 17 |
| Not Dead nor censored | 3 | 0 | 3 |
| **MGMT Promoter Status** | | | |
| Hypermethylated | 23 | 5 | 18 |
| Unmethylated | 33 | 6 | 27 |
| Indeterminate | 14 | 2 | 12 |
| **IDH-1 Mutation Status** | | | |
| Wild-Type | 59 | 12 | 47 |
| Mutant | 8 | 1 | 7 |
| Indeterminate | 3 | 0 | 3 |
| **1p-19q-Codeletion Status** | | | |
| Codeletion | 2 | 0 | 2 |
| Negative | 18 | 1 | 17 |
| Indeterminate | 50 | 12 | 38 |

\* Some of the censored patients are expired (dead), therefore all the censored patients are not alive.

*2.2 Algorithm pipeline for prediction of rapid early progression (REP)*

The overall pipeline for prediction of rapid early progression (REP) is illustrated in Figure 1. Following the pipeline, selected features and algorithm are depicted in Table 2, Table 3 and Fig. 2 respectively. The details of each of the steps are described in the following sub-sections. The prediction model is ensemble tree-based Cat Boost Classification model. The performance of the model is evaluated with 1000 iterations and 5-fold cross-validation which yield 5000 different values in Area Under Curve (AUC), Positive Predictive Value (PPV), False Positive Rate (FPR), and Accuracy. The statistical distribution of the 5000 values is considered to describe model performance.

*2.3 MRI Preprocessing, Tumor Volume Segmentation and Feature Extraction*

2.3.1 MRI Preprocessing

All radiation-planning MRI images are co-registered to the same T1 anatomic template using affine registration and resampled to $1mm^3$ voxel resolution using the Oxford Center for Functional MRI of the Brain's (FM- RIB) Linear Image Registration Tool (FLIRT) of FMRIB Software Library (FSL) [20]. The FSL's Brain Extraction Tool (BET) [21] is used to skull-strip each patient's volumetric image. To obtain improved skull-stripped volumetric images, manual intervention is used. Additionally, all images are smoothed using the Smallest Unvalued Segment Assimilating Nucleus (SUSAN) [22], a low-level image processing technique, to reduce high frequency intensity changes (i.e., noise) in regions with uniform intensity profile while maintaining the underlying structure. The intensity histograms of all modalities for all patients are then matched to the relevant modality of a single reference patient using the implemented version of Insight Toolkit( ITK ) [23].

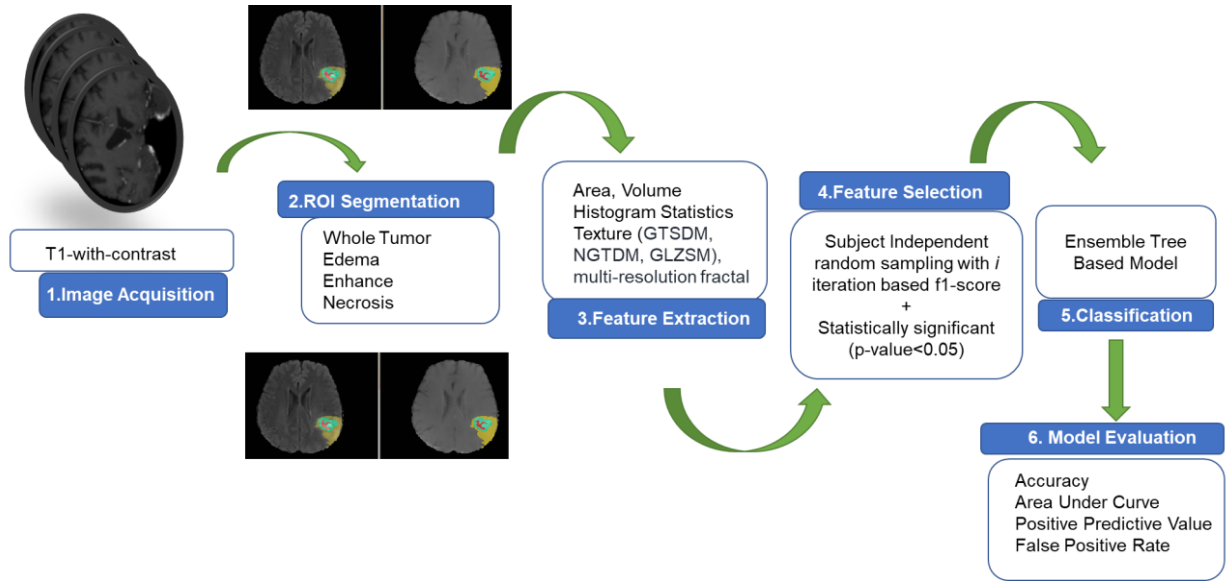

**Figure 1.** Overall pipeline for rapid early progression prediction.

2.3.2 Tumor Volume Segmentation

Due to limited patient data, transfer learning is used for tumor volume segmentation. For training of the segmentation model, we utilized 1251 GB patient cases collected from BRATS 2021 challenge [24–26]. Then, the trained model is utilized to segment the tumor tissue region in the 70 patient cohort with radiation planning MRI volumetric images. For segmentation of tumor tissue regions (edema, enhancing tumor and necrosis), we trained a 3D [13,27] UNet model with T1C sequence of MRI scans of glioblastoma patients. The segmented tumor regions are verified by radiation oncologists.

2.3.3 Feature Extraction

About 600 features are extracted from the tumor tissue volume segments in this study. These features include texture, volume, and the area of the tumor and its sub-regions (edema, enhancing tumor, and necrosis). Forty-one texture characteristics are recovered from the whole tumor volume in raw MRI (T1C) sequence, as well as tumor sub-regions. The conventional texture features are extracted from grey-tone spatial dependence matrices (GTSDM), neighborhood grey-tone difference matrix (NGTDM), and gray level size zone matrix (GLZSM). The fractal texture features are piecewise triangular prism surface area (PTPSA) for fractal characterization, multi-resolution Brownian motion (mBm) analysis, and tumor region characterization with Holder Exponent (HE) modeling termed as generalized multi-resolution Brownian motion (GmBm) [28–30]. Multiresolution fractal features describe textural variation in tumor tissue across many image resolutions [31]. Six histogram-based statistics (mean, variance, skewness, kurtosis, energy, and entropy) are also derived from the distinct tumor sub-regions. We further extract volumetric features: the volume of the entire tumor, the volume of the whole tumor in relation to the brain, and the volume of sub-regions.

2.4 Radiomic Features Selection and Model Building

To evaluate the efficacy of fractal features and conventional radiomics features, we consider two model configurations. These model configurations are (a) non-fractal model (model with only conventional volume, area, and texture features), (b) fractal model (model with multi-resolution fractal features and conventional volume, area, and texture features)[32]. Radiomic modeling of REP classification using the extracted features is implemented using a nested paradigm of 1000 iterations and 5-fold cross-validation with random sampling. The group $D_P$ indicates the data frame of 13 patients with REP class (denoted as 1) and $D_N$ denotes non-REP (denoted as 0). Since there is class size imbalance, a subsample of group $D_N$ is selected, repetitively under a bootstrap algorithm and compared with group $D_P$. For each $i$ iteration, 20 patients are randomly sampled from 57 non-REP cases. The reason to select 20 samples is to have around 40-50% cases from REP group and the rest from non-REP group in each $i$ iteration. Under each $i$ iteration there is n-fold cross validation, denoted as *n* in Fig. 2. Under each n-fold there is feature matrix $\{X_j\}_{j=1\ to\ k}$, where *k* denotes 600 extracted features. The F1-score represents how well a feature performs in categorizing the positive (minority) data, making it an important indicator when the underlying data distribution is unbalanced [33]. After completing *i* iterations, the mean F1-score is calculated for each of the 600 features. From our analysis, the range of mean F1-score for the 600 features is 0.0-0.74. The features score greater or equal to 0.6 (threshold value) is chosen. The reason to select 0.6 as the threshold value close to 1.00 (the highest value of F1-score) the better. There are seven features after applying thresholding in non-fractal and fractal models. In the second step, the statistically significant (p-value<0.05) feature among the selected seven features is depicted in Table 2 for non-fractal model, and in Table 3 for fractal model, respectively. The overall algorithm for subject independent random sampling of the first step feature selection process is illustrated in Fig. 2. A detailed statistical analysis of selected features is presented in the results section.

The model for REP prediction follows the *n*-fold cross validation with *i* iteration as illustrated in Figure 3. The selected three features for each of the non-fractal and fractal models are used to evaluate the respective model performance. The model performance evaluation metric *P* denotes the accuracy, AUC, PPV and FPR while *R* denotes the ranking score of features. Finally, the mean and standard deviation deviations of performance metrics after *i* iterations are reported.

---

Algorithm: Subject Independent random sampling with *i* iteration for feature ranking

---

*Input*: iteration number *i*, number of folds *n*, Radiomics feature data frame *D*, data frame with REP status $D_P$, data frame with non-REP status $D_N$

*Define* $S_N$ as 20 patients randomly sampled from 57 non-REP $D_N$ in each iteration *i*, $S_P$ 13 patients with REP status in each iteration *i*

*Define Y* as target class vector and $\{X_j\}_{j=1\ to\ k}$ the feature matrix in the data frame D, *k* is number of total features.

for **iteration= 1** to **i** do

    **Initialize** $S_N \longleftarrow D_N$, randomly sample 20 patients from 57 non-REP patients.

    **Initialize** $S_P = D_P$

    **Initialize** $D = \{D_N, D_P\}$

    **Save** *D* after each $i^{th}$ iteration

    within *D* split the target variable vector *y* and feature variable matrix as $\{X_j\}_{j=1\ to\ k}$

        for **fold= 1** to **n** do

            enumerate train and test index for n-fold in $\{X_j\}_{j=1\ to\ k}$ and *y*

            for **j= 1** to **k** do

            fit a decision tree classifier (DT) on the train index of $\{X_j\}$

                predict $\hat{y}$ on the test index of $\{X_j\}$

                calculate F1-score based on true *y* and predicted $\hat{y}$ of the test index of $\{X_j\}$

                assign the score as the feature score of each $\{X_j\}$

            end for

        end for

        **Output:** Cumulative f1-score after n-fold validation

end for

**Output:** Rank features based on cumulative score after *i* iterations.

---

**Figure 2.** Algorithm for subject independent random sampling with *i* iteration for first step feature selection.

**Table 2.** The final three significant features selected in non-fractal model for REP prediction.

| Original Extracted Features for Non-Fractal Model | Significant (p-value<0.05) features from second step feature selection (Rank score) |
|---|---|
| 41 texture features extracted from raw T1C | • Autocorrelation of grey-tone spatial dependence matrices (GTSDM) from T1C (0.677) |
| 12 volumetric features | None |
| 9 area related features | • Eccentricity in edema region (0.654) <br> • Second axis (y-axis) length in necrosis region (0.744) |
| 6 Histogram Statistics | None |

**Table 3.** The final three significant selected features in fractal model for REP prediction.

| Original Extracted Features for Fractal Model | Significant (p-value<0.05) features from second step feature selection (Rank score) |
|---|---|
| 41 texture features extracted from PTPSA, mBm and GmBm of T1C (fractal features) | • Generalized multi-fractal Brownian motion (GmBm) of T1C (0.744) |
| 41 texture features extracted from raw T1C | • strength of NGTDM from 37th direction on basis of T1C image size (0.642) <br> • strength of NGTDM from raw T1C (0.648) |
| 12 volumetric features | None |
| 9 area related features | None |
| 6 Histogram Statistics | None |

---

Algorithm: Subject independent *n*-fold cross validation with *i* iteration for model evaluation

*Input*: D after each *i* iterations, ranking score R for each feature in matrix $\{X_j\}_{j=1\ to\ k}$

*Define* $\{X_S\}$ selected feature matrix based on R which is a subset of $\{X_j\}_{j=1\ to\ k}$, Classification model C, Model performance evaluation metrics P

*for* **iteration= 1** *to* ***i*** *do*

    **Initialize** $D = \{D_N, D_P\}$

    **Initialize** $\{X_S\}$ based on R and target vector y from D

        *for* **fold= 1** *to* ***n*** *do*

            enumerate train and test index for n-fold in $\{X_S\}$ and *y*

            fit classifier model C to the train index of $\{X_S\}$ and *y*

            evaluate C on the test index of $\{X_S\}$

            Save P after each *n*

    *end for*

        **Output**: Cumulative P after *n* fold

*end for*

**Output:** Mean values of *P* after *i* iterations.

**Figure 3.** Algorithm for subject independent n-fold cross validation with *i* iteration for model evaluation.

*2.5 Survival Analysis Modeling Under Dependent Censoring*

The Kaplan-Meier estimator and Cox proportional hazard model are typical methods for survival analysis and feature selection [17] techniques that manage with censoring under the presumption that overall survival time and censoring time are statistically independent. Therefore, a copula-based approach [17] is used to estimate the dependence parameter by utilizing cross validation to select significant gene or features for survival prediction. The dependency between survival and censoring time is modeled by copula functions.

Considering random variables, if T is the survival time and U is the censoring time and $X_i = (X_{i1}, X_{i2}, ..., X_{ip})$ is the p vectors of features [16], the response or the target is $(t_i, \delta_i, X_i)$ where $t_i = \min\{T,U\}$ and $\delta_i = I\{T_i \leq U_i\}$, where I {.} is the indicator function which indicates whether the time is survival time or censoring time. The univariate cox hazard model [34] is represented as,

$$h(t|x_{ij}) = h_o(t) e^{\beta_{ij} x_{ij}}, \quad j = 1, 2, \ldots p, \tag{1}$$

where, p indicates the number of predictors. The estimator is $\widehat{\beta_j}$ is used to obtain p-value from the Wald test under the null hypothesis $H_{oj}: \beta_j = 0$. The features or genes that are selected have p-values under a certain threshold level of significance value. The estimator $\widehat{\beta_j}$ can correctly estimate $\beta_j$ if the independence assumption is satisfied [35].

A copula model for dependent censoring [36] is presented as,

$$\Pr(T > t, U > u) = C_a(S_T(t), S_U(u)), \tag{2}$$

where, $C_a$ is a bivariate copula function and $S_T(t) = \Pr(T > t)$ and $S_U(u) = \Pr(U > u)$ indicate marginal survival. Under independence assumption, $\alpha = 0$ which leads to $C_\alpha(u, v) = uv$, [36] the resultant p-values same as univariate cox model p-values and the equation (2) under independence assumption can be written as the following:

$$\Pr(T > t, U > u) = \Pr(T > t) \Pr(U > u). \tag{3}$$

However, there are bivariate copula functions that may be considered for the potential correlation between the censoring and the survival times. We select the Clayton copula because of its mathematical simplicity [16]. It is suitable for statistical analysis because of positive dependence structure property and captured by parameter alpha ($\alpha$) described in Equation 2. Moreover, the Clayton copula may be particularly applicable in cancer survival because it concentrates on the reliance in the lower tail of the bivariate density function, which represents the prevalent circumstance in which a rapid time to progression leads to a rapid time to death [37].

Under dependence censoring [36], the Clayton copula $C_\alpha(u, v)$ can be represented as the following,

$$C_\alpha(u, v) = (u^{-\alpha} + v^{-\alpha} - 1)^{-\frac{1}{\alpha}}, \alpha > 0. \tag{4}$$

The parameter $\alpha$ depicts degree of dependence [17] and can be converted to Kendall's tau ($\tau$), under Clayton copula the tau is $\tau = \alpha/(\alpha + 2)$. Under the assumption, all the observed times are unique ($t_i \neq t_j$, where $i \neq j$) **among patients** for each patient. The number of at-risk at time $t_i$, $n_i = \sum_{l=1}^{n} I(t_l \geq t_i)$, under Clayton copula, has-graphic estimator [36] is presented as,

$$\widehat{S_T}(t) = \left[1 + \sum_{t_i \leq t, \delta_i = 1} \left\{\left(\frac{n_i - 1}{n}\right)^{-\alpha} - \left(\frac{n_i}{n}\right)^{-\alpha}\right\}\right]^{-1/\alpha}, \tag{5}$$

The copula-graphic estimator is equivalent to the Kaplan-Meier estimator under the independence copula [16]. Given survival data $\{(t_i, \delta_i, x_{ij}); i = 1, \ldots, n\}$ for j-th feature (j=1,2,…p), the data is modeled using the following copula equation.

$$\Pr(T > t, U > u | x_j) = C_\alpha\{\Pr(T > t | x_j), \Pr(U > u | x_j)\}, \tag{6}$$

where $\Pr(T > t|x_j) = \exp\{-\wedge_{0j}(t)\exp(\beta_j x_j)\}$, $\Pr(U > u|x_j) = \exp\{-\Gamma_{oj}(u)\exp(Y_j x_j)\}$, and the same copula $C_\alpha$ for every $j$. Here, $\beta_j, Y_j, \beta_j=(\beta_{oj}, \beta_{1j},\ldots,\beta_{pj})$; are the regression co-efficient and $\wedge_{0j}, \Gamma_{oj}$ are the cumulative baseline hazard function which are unspecified [16]. For $\alpha$, the semiparametric maximum likelihood estimator ($\widehat{\beta_j}(\alpha)$, $\widehat{Y_j}(\alpha)$, $\widehat{\wedge_{oj}}(\alpha)$, $\widehat{\Gamma_{oj}}(\alpha)$) determined using dependCox.reg() function in R [17]. Harrell's concordance c-index [38,39] is a robust predictive measure to calculate $\alpha$. Therefore, $\alpha$ is selected by maximizing cross-validated c-index. The c-index is a measure of concordance between the outcome or target $(t_i, \delta_i)$ and the predictors $\sum_{j\in\Omega}\widehat{\beta_j^{(-i)}}(\alpha)x_j$. For a given $x_j$, the bivariate survival function [16] is the following:

$$\Pr(T > t, U > u|x_j) = \phi_{\beta(-j),Y(-j)}[\phi_{\beta(-j)}^{-1}\{\Pr(T > t|x_j)\}, \phi_{Y(-j)}^{-1}\{\Pr(U > u|x_j)\}], \quad (7)$$

where, $\phi_{\beta(-j),Y(-j)}(u,v)=E\{\exp(-uX_j-vY_j)|x_j\}$, $\phi_{\beta(-j)}(u) = \phi_{\beta(-j),Y(-j)}(u,0)$, and $\phi_{Y(-j)}(v)=\phi_{\beta(-j),Y(-j)}(0,v)$ are Laplace transforms. For $\beta=Y$, Eq. (7) is called the Archimedean copula model and is represented as follows to model the survival model under censoring:

$$\Pr(T > t, U > u|x_j) = \phi_{\beta(-j)}[\phi_{\beta(-j)}^{-1}\{\Pr(T > t|x_j)\} + \phi_{\beta(-j)}^{-1}\{\Pr(U > u|x_j)\}]. \quad (8)$$

The dependency between T and U for a given $x_j$, is described as in Equation (6) under Sklar's theorem [16,40]. copula reduces the restrictive independent condition of $C_\alpha(u,v) = uv$. Using dependCox.reg.CV( )[35] is utilized to determine $\widehat{\beta_1}(\widehat{\alpha}),\ldots,\widehat{\beta_p}(\widehat{\alpha})$.

*2.6 Survival Prediction*

Figure 4 represents the overall pipeline for OS prediction. The goal is prediction of patients with expired/dead status. The copula-based method is computationally expensive [36], therefore, we apply a first step feature selection using the algorithm in Figure 2. Afterwards, based on F1-score range (0.2-0.84), the features F1-score greater than 0.7 are selected. The selected number of features is 124 and 87, after the first step in feature selection for non-fractal and fractal model, respectively. The selected features are used as input for a second step feature selection using the copula model to identify significant (p-value<0.05) features. These final selected features are related to patients' OS probability. In addition, we also analyze whether the inclusion of molecular features (MGMT status, IDH status) is significantly related to patient survival probability.

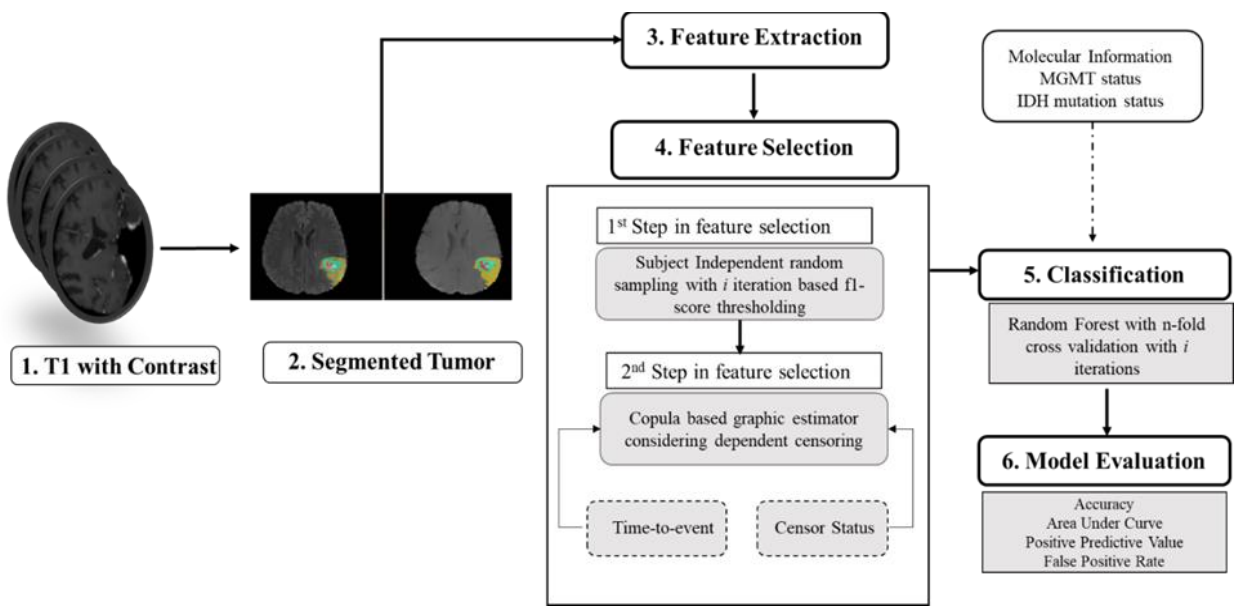

**Figure 4.** Overall pipeline for prediction of survival.

Table 4. Stratification of patient's censored and expiration status

|  | Expired/Dead (denoted as 1) | Alive (denoted as 0) |
|---|---|---|
| Non-Censored Patient (n=45) | 45 | None |
| Censored/Lost Follow-up Patient (n=22) | 9 | 13 |

Considering the censoring scenario, we propose to apply copula method for feature selection in survivability analysis. At first, we identify the dependency parameter α utilizing the survival data $(t_i, \delta_i)$ and the extracted radiomics features matrix $X_i = (X_{i1}, X_{i2}, ..., X_{ip})$. In addition to imaging features, molecular features are also added in the feature matrix. The significant (p-value <0.05) features are then utilized to make binary prediction whether the patient is expired/dead or alive.

## 3. Results

### 3.1. Prediction performance of Rapid Early Progression (REP) Classification

The comparative performance of the two model configurations is illustrated in Table 5. When comparing the performance of the fractal to non-fractal model, the fractal model achieves an AUC of 0.793 while the non-fractal model achieves an AUC of 0.673. There is significant difference (ANOVA test, p-value<0.001) in accuracy, AUC, PPV and FPR between two model configurations as presented in Figure 5.

Table 5. Comparison between model configurations for 5-fold cross validation over 1000 iterations with subject independent random sampling for REP classification

| Model Configurations | Area Under Curve (AUC) | Accuracy (%) | Positive Predictive Value (PPV) | False Positive Rate (FPR) |
|---|---|---|---|---|
| **Non-Fractal Model** | 0.673 ± 0.082 | 63.5 ± 0.069 | 0.617 ± 0.067 | 0.262 ± 0.177 |
| **Fractal Model** | 0.793 ± 0.082 | 78.1 ± 0.071 | 0.761± 0.069 | 0.145 ± 0.107 |

The statistically significant features in non-fractal model are eccentricity in edema region, second axis (y-axis) length in necrosis region and autocorrelation of grey-tone spatial dependence matrices (GTSDM) from T1C, respectively. To identify the significant difference of features between REP and non-REP groups, we first observe the distribution of features in each group. The Shapiro-Wilk test is performed to analyze the underlying distribution of each feature vector. In case of eccentricity in edema region the distribution is not normal therefore Wilcoxon-Mann-Whitney test is performed. However, for the other two features the distribution is normal and ANOVA test is performed. For statistical analysis, p-value <=0.05 is considered significant. From Table 6, we observe the significant difference of the features between the two groups. For instance, the median and mean value of REP group in necrosis region is higher than non-REP; similar trend is observed in autocorrelation of GTSDM from T1C between two groups.

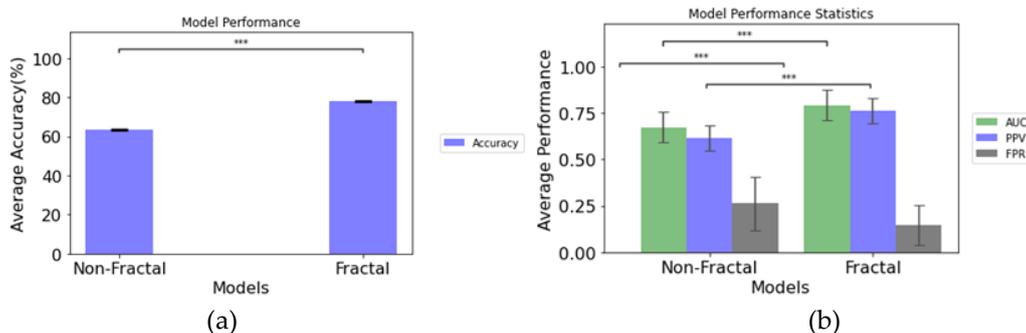

(a) (b)

**Figure 5.** (a) Accuracy of Non-Fractal and Fractal Model, (b) Performance statistics of Non-fractal and Fractal Model. Error bar represents standard deviation and * depicts significant differences between the model configurations.

**Table 6.** Descriptive statistics of selected features in Non-fractal Model

|  | Non-REP (n=57) | REP (n=13) | p-value |
|---|---|---|---|
| **Eccentricity in edema region** |  |  | 0.0446 |
| Mean(±std) | 0.8320±0.1771 | 0.7779±0.1184 |  |
| Standard error | 0.0235 | 0.0328 |  |
| Median | 0.8922 | 0.8109 |  |
| **Second axis (y-axis) length in necrosis region** |  |  | 0.0116 |
| Mean(±std) | 0.5445±0.1995 | 0.7072±0.2239 |  |
| Standard error | 0.0264 | 0.0621 |  |
| Median | 0.5597 | 0.8085 |  |
| **Autocorrelation of grey-tone spatial dependence matrices (GTSDM) from T1C** |  |  | 0.0262 |
| Mean(±std) | 0.3731±0.2125 | 0.5137±0.1381 |  |
| Standard error | 0.0281 | 0.0383 |  |
| Median | 0.3522 | 0.4984 |  |

For the fractal model, the statistically significant features are the generalized multi-resolution Brownian motion (GmBm) of neighborhood grey-tone difference matrix (NGTDM) of T1C, strength of NGTDM from 37$^{th}$ direction of T1C and strength of NGTDM. We observe the feature distribution is not normal in each group and therefore Wilcoxon-Mann-Whitney test is performed for significant difference (p-value<0.05) of feature distribution between two groups. In fractal model, the significant selected features are texture features. The median value of the selected features is higher in non-REP group compared to REP as shown in Table 7.

**Table 7.** Descriptive Statistics of selected features in Fractal Model

|  | Non-REP (n=57) | REP (n=13) | p-value |
|---|---|---|---|
| **GmBm* of T1C** |  |  | 0.0296 |
| Mean(±std) | 0.3587±0.2012 | 0.2372±0.1272 |  |
| Standard error | 0.0267 | 0.0353 |  |
| Median | 0.3325 | 0.1918 |  |
| **Strength of NGTDM from 37$^{th}$ direction of T1C** |  |  | 0.0019 |
| Mean(±std) | 1.9390±0.5780 | 0.8260±0.4864 |  |

|  |  |  |
|---|---|---|
| Standard error | 0.0415 | 0.1349 |
| Median | 1.2292 | 0.6970 |
| **Strength of NGTDM** |  | 0.0013 |
| Mean(±std) | 0.1224±0.0665 | 0.0389±0.0299 |
| Standard error | 0.0221 | 0.0083 |
| Median | 0.0752 | 0.0252 |

*GmBm: generalized multi-resolution Brownian motion

*3.3. Survival probability analysis under dependent censoring*

First, we analyze the impact of dependent censoring in feature selection for survival probability. For this purpose, we evaluate the significant (p-value<0.05) features using Cox proportional hazard model with independent censoring. The features selected with independent censoring are compared with features selected with dependent censoring. Table 8 represents the selected significant (p-value<0.05) features with independent censoring using Cox proportional model. Table 9 represents the significant features (p-value<0.05) with dependent censoring utilizing copula modeling. In the case of non-fractal models, no significant features are selected applying Cox modeling. To examine the effect of dependent censoring in feature selection, we compare the survival probability curve utilizing the top 2 features in fractal model. As, with independent censoring only 2 features are selected, therefore, we analyze the survival marginal curves with and without censoring dependency.

**Table 8.** Significant features using Cox Proportional Hazard Model (Independent Censoring)

| Features Name | Coefficient | p-value |
|---|---|---|
| T1C_mBm_GLZSM_LargeZoneLowGrayEmphasis[1] | 5.46 | 0.002 |
| T1C_ptpsa_GLZSM_LargeZoneLowGrayEmphasis[2] | 2.46 | 0.038 |

[1] multi-resolution Brownian motion (mBm), gray level size zone matrix (GLZSM) of T1C MRI sequence
[2] piecewise triangular prism surface area (ptpsa) of T1C MRI sequence

Table 9. Significant features using copula modeling (Dependent Censoring). The features are ordered according to p-value.

**Fractal Model Features**

| Features Name | Co-efficient | p-value |
|---|---|---|
| ET2[1] | -1.58 | 0.0045 |
| T1C_ptpsa_GLZSM_Low_Gray_Level_Zone_Emphasis | 1.33 | 0.0110 |
| L2_Orientation[2] | 0.74 | 0.0183 |
| edema_FirstAxisLength[3] | -0.91 | 0.0194 |
| wt_MajorAxisLength[4] | -0.93 | 0.0198 |
| L1_Extent[2] | 0.76 | 0.0218 |
| L3_Orientation[2] | 0.70 | 0.0261 |
| T1C_mBm_GLZSM_LargeZoneLowGrayEmphasis | 2.27 | 0.0316 |
| nec_SecondAxis_1[5] | -1.06 | 0.0355 |
| T1C_ED_Histogram_Mean[6] | 1.09 | 0.0434 |

**Non-fractal Model Features**

| Features Name | Co-efficient | p-value |
|---|---|---|
| ET2[1] | -1.58 | 0.0045 |
| L2_Orientation[2] | 0.74 | 0.0183 |
| edema_FirstAxisLength[3] | -0.91 | 0.0194 |
| wt_MajorAxisLength[4] | -0.93 | 0.0198 |
| L1_Extent[2] | 0.76 | 0.0218 |
| L3_Orientation[2] | 0.70 | 0.0261 |
| nec_SecondAxis_1 | -1.06 | 0.0355 |
| T1C_ED_Histogram_Mean[6] | 1.09 | 0.0434 |
| ED_up_left_y* | 0.83 | 0.0585 |
| T1C_ED_Histogram_Skewness* | -1.18 | 0.0718 |

*Features are not significant, only 8 features are significant in non-fractal configurations
[1]Eccentricity of whole tumor region; [2] L1, L2, L3 indicates x, y, z axis of whole tumor region.
[3]major or first axis length from edema region, [4] major or first axis length of whole tumor region.
[5]second or y-axis length of necrotic region, [6] histogram statistics of edema region with T1C sequence.

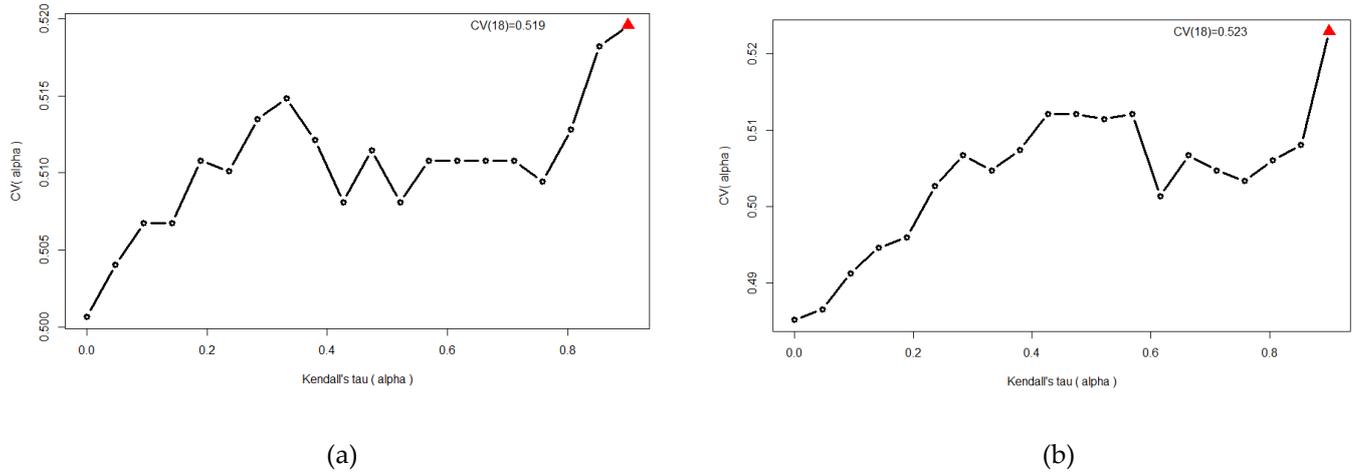

**Figure 6.** The cross validated c-index for the fractal and non-fractal model. The cross validated c-index is maximized at α=18 which corresponds to Kendall's tau=0.90. (a) c-index= 0.519 for fractal model, (b) c-index=0.523 for non-fractal model.

From Table 8, we observe the two selected features by Cox model under independent censoring assumptions. Both features refer to fractal texture features extracted from MRI. For a patient case with feature vector $x = (x_1, x_2, ...., x_p)'$, survival prediction is analyzed using prognostic index (PI) defined as $\widehat{\beta(\alpha)}'x$, where, $\widehat{\beta(\alpha)}' = (\widehat{\beta_1}(\alpha),...., \widehat{\beta_p}(\alpha)$ [16,35]. If α=0, then PI=$\widehat{\beta(0)}'x$ which is based on Cox modeling under independent censoring (α=0). Therefore, the PI for the fractal modeling with cox model with two significant features is the following.

$$\text{PI (with independent censoring)} = (5.46*\text{T1C\_mBm\_GLZSM\_LargeZoneLowGrayEmphasis}) \\ +(2.46* \text{T1C\_ptpsa\_GLZSM\_LargeZoneLowGrayEmphasis}). \quad (9)$$

However, considering dependent censoring (α=18), utilizing the copula modeling the top 2 selected features as shown in Table 9 and the prognostic index for fractal model is the following:

$$\text{PI (with dependent censoring)} = (-1.58*\text{ET2}) + (0.74* \text{L2\_Orientation}). \quad (10)$$

Using the PI, we randomly divide the 67 patient cases into two groups of equal sample size ($n_1$=33, $n_2$=34). Patients in good(low-risk) prognostic group with low PIs and patients in bad (high-risk) prognostic group with high PIs [16,17,36]. The two survival curves are determined by copula graphic (CG) estimator [41,42] with Clayton copula. The difference between two curves is calculated by average vertical difference [16,17,43]. The p-value is calculated between two groups using 1000 permutation tests [16,17,44]. From Figure 7(a), we observe that the vertical distance (D=0.128) between two groups with independent censoring (α=0), is not significant (p-value=0.1422). However, considering the dependent censoring in Figure 7(b) (α=18, c-index= 0.519), the distance (D=0.185) between two prognostic groups is significant (p-value=0.0047).

*3.4. Binary prediction of survival*

From the analysis, we observe the effect of dependent censoring on feature selection and survival probability. Therefore, for binary survival prediction we utilize the selected features by copula modeling. Table 9 presents the significant (p-value<0.05) features for fractal and non-fractal models. For experimental analysis of selected features, we consider the top 3,5,7, and 9 features for binary classification whether the patient is expired/dead or not. In the case of fractal model, 10 features are significant for survival probability analysis while that for non-fractal model 8 features are significant as presented in Table 9. First, we analyze the vertical distance between good prognostic and bad prognostic groups with 3,5,7,9, and 10 features. Afterwards, based on the significance of feature combination we compute binary prediction of survival.

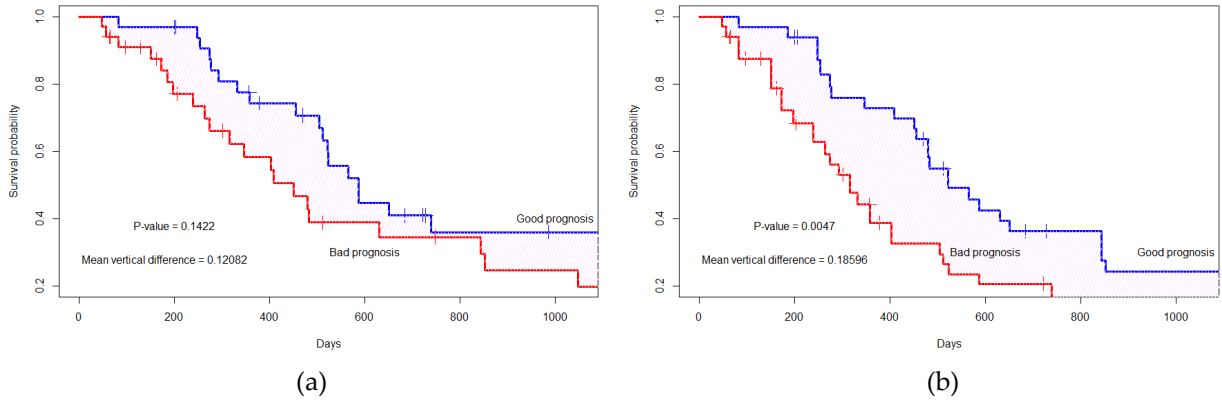

**Figure 7**. The marginal survival curve is separated by top 2 significant features considering (a) independent censoring (where $\alpha=0$), p-value= 0.1422, (b) dependent censoring (where $\alpha=18$, c-index=0.519), p-value= 0.0047. The '+' indicates the censored cases.

Based on the selected top 3,5,7, and 9 features we use the algorithm in Figure 3 for binary prediction of survival. The predictive results for 5-fold cross validation with 1000 iterations are presented in Table 10. In the case of binary prediction of survival, we focus on the precision of the models to selected feature numbers. Because with higher precision we have a low false positive rate as shown in Table 10. Moreover, following the algorithm in Figure 3, we consider balanced number of expired/dead (n=25, 25 cases randomly sampled from 54 dead cases) and not dead (n=13) cases in each iteration. It is observed from Figure 8 that when utilizing seven features, both model configurations (fractal, non-fractal) achieve a higher PPV or precision. Therefore, we compare the model configurations performance based on the top seven features models. While comparing the fractal and non-fractal model configuration performance for binary survival prediction, there is significant difference in area under curve (AUC) with p-value=0.005 from ANOVA analysis. Moreover, in terms of PPV (p-value<0.01), accuracy (p-value<0.001), false positive rate (p-value<0.01) there is significant difference between the performance of fractal and non-fractal model.

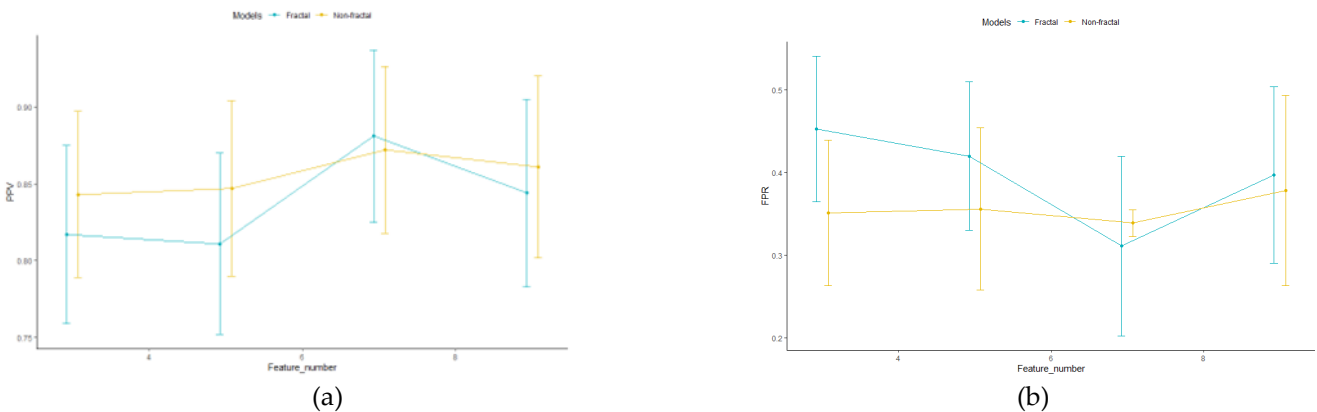

**Figure 8**. The fractal, non-fractal model configurations positive predictive value (a) PPV, (b)FPR, for 5-fold 1000 iterations for different number of features. Points indicate the mean PPV over 5-fold 1000 iterations. The error bar indicates the standard deviation.

Table 10. Comparison across model configurations of mean test performance across 5-fold with 1000 iterations for binary prediction of patient survival (patient expired(dead) or not).

| Number Of features | Model Configurations | Area Under Curve (AUC) | Accuracy (%) | Positive Predicted Value (PPV) | False Positive Rate (FPR) |
|---|---|---|---|---|---|
| Top 3 features | Non-fractal Model | 0.730±0.235 | 74.018±0.045 | 0.843±0.054 | 0.351±0.088 |
| | Fractal Model | 0.659±0.241 | 67.67±0.040 | 0.817±0.058 | 0.453±0.088 |
| Top 5 features | Non-fractal Model | 0.783±0.199 | 73.22±0.048 | 0.847±0.057 | 0.356±0.098 |
| | Fractal Model | 0.658±0.243 | 70.03±0.048 | 0.811±0.059 | 0.420±0.090 |
| Top 7 features | Non-fractal Model | 0.735±0.219 | 73.05±0.045 | 0.872±0.054 | 0.339±0.106 |
| | Fractal Model | 0.762±0.214 | 74.39±0.046 | **0.881±0.056** | **0.311±0.109** |
| Top 9 features | Non-fractal Model | 0.725±0.224 | 71.00±0.049 | 0.861±0.059 | 0.378±0.115 |
| | Fractal Model | 0.719±0.214 | 70.37±0.048 | 0.844±0.061 | 0.397±0.107 |

In addition to radiomics features, we analyze the survival probability and binary prediction of survival with molecular information (MGMT methylation status, IDH status). Therefore, with the top 7 selected features we included MGMT status and IDH mutation status as additional features. In both fractal and non-fractal models the MGMT status is not significant (p-value=0.9651) under dependent censoring copula modeling. However, IDH mutation status is significant (p-value =0.04) and therefore, we added IDH mutation status with the top 7 features and compute model performance as presented in Table 11. With additional molecular features, the distance between marginal survival curves is almost the same. The vertical distance D= 0.171 with p-value=0.0085 as shown in Figure 9 for fractal model. In non-fractal model, the vertical distance D=0.173 with p-value =0.0084.

Table 11. Comparison across model configurations (top 7 features with molecular status (IDH mutation) of mean test performance across 5-fold with 1000 iterations for binary prediction of patient survival (patient expired(dead) or not).

| Model Configurations | Area Under Curve | Accuracy (%) | Positive Predictive Value (PPV) | False Positive Rate (FPR) |
|---|---|---|---|---|
| Non-Fractal-Molecular | 0.757±0.214 | 72.36 ±0.046 | 0.866±0.058 | 0.354±0.109 |
| Fractal Molecular | 0.762±0.214 | 73.48±0.047 | 0.883±0.057 | 0.322±0.114 |

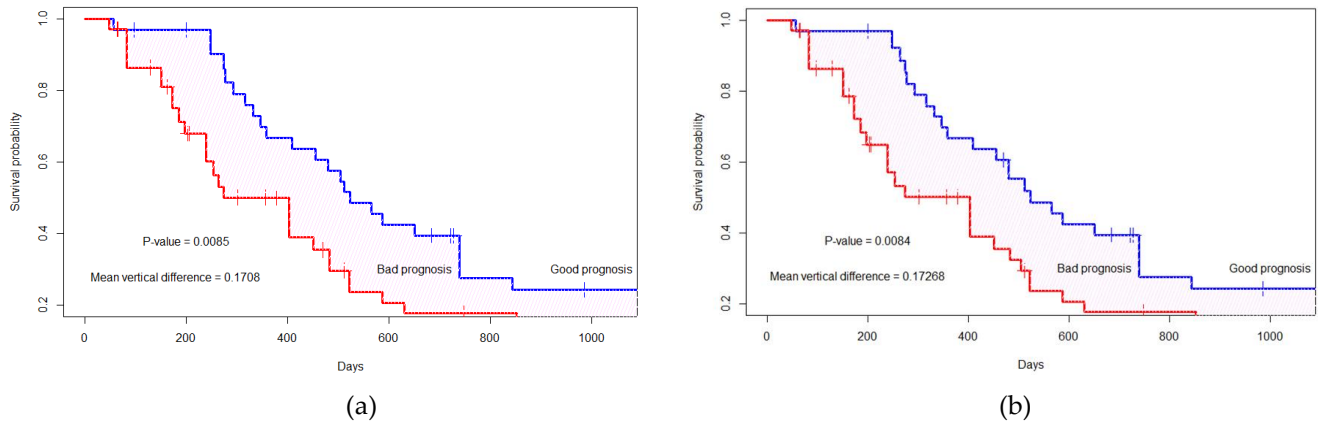

**Figure 9**. The marginal survival curve for good (or bad) prognosis group is separated by top 7 features+ molecular status (IDH mutation). (a) Fractal-Molecular Model, (b) Non-fractal Molecular Model. The '+' indicates the censored cases.

For fractal-molecular and non-fractal-molecular model there is a significant difference between accuracy, PPV and FPR. However, there is no significant difference between the model configurations in terms of AUC. The reason could be as shown in Figure 9 in terms of survival probability the vertical distance and p-value is almost same for both model with addition of molecular features with 7 radiomics features. Moreover, with inclusion of molecular features, there is no increment in model performance in terms of PPV, FPR compared to model configurations without molecular feature.

*3.5. Analysis of Prognostic Groups and its association with REP status*

From Table 9 we observe that in case of the non-fractal model a total of 8 features are significant (p-value<0.05) while that for the fractal model is 10, respectively. To analyze the distribution of REP patients in prognostic (good or bad) groups, we consider the top 8 features in both fractal and non-fractal models. Therefore, the PI for fractal and non-fractal models are the following,

$$\text{PI (Fractal model)} = (-1.58*ET2)+(1.33*T1C\_ptpsa\_GLZSM\_LGLZE)+ (0.74*L2\_Orientation)+(-0.91*edema\_First\_Axis\_Length) + (-0.93*wt\_Major\_Axis\_Length)+ (0.76*L1\_Extent) + (0.70*L3\_Orientation) + (2.27*T1C\_mBm\_GLZSM\_LZLGE),$$ (11)

and

$$\text{PI (Non-fractal model)} = (-1.58*ET2) + (0.74*L2\_Orientation) +(-0.91*edema\_First\_Axis\_Length) + (-0.93*wt\_Major\_Axis\_Length)+ (0.76*L1\_Extent) + (0.70*L3\_Orientation) + (-1.06*nec\_Second\_Axis\_1).$$ (12)

The PI is calculated using the extracted radiomics and multi-resolution fractal features from radiation planning MRI. The only difference between the PI of fractal and non-fractal models is that in fractal models we include the multiresolution fractal texture features in addition to conventional texture features. Sixty-seven patients are divided into good prognostic and bad prognostic groups. Using copula graphic estimator two marginal survival curves are determined. In fractal model the distance between the two marginal curves is D=0.157 (p-value=0.014) while that non-fractal model is D=0.128 (p-value=0.038), respectively.

Figure 10 shows the distribution of REP cases in the prognostic groups. The scatter dots represent the prognostic index for each patient. The darker point indicates the mean PI under each group and the bar represents the standard deviation. For instance, in Figure 10 (a), within the scatter plot the group "c" indicates the patients in bad prognostic group and, they are also rapid early progression cases. From Figure 10, we observe that patients with low prognostic index (PI)/lower risk are in the good prognostic group. The patients with higher prognostic index (PI)/higher risk are in the bad prognostic group. In addition, in the case of fractal model, based on prognostic index 84.62 % (11 out of 13 cases) of REP cases fall under bad prognostic group as shown in the matrix representation of Figure 10(a). In non-fractal model 76.92 % (10 out of 13 cases) of REP cases belongs to bad prognostic group as presented in Figure 10(b) matrix representation.

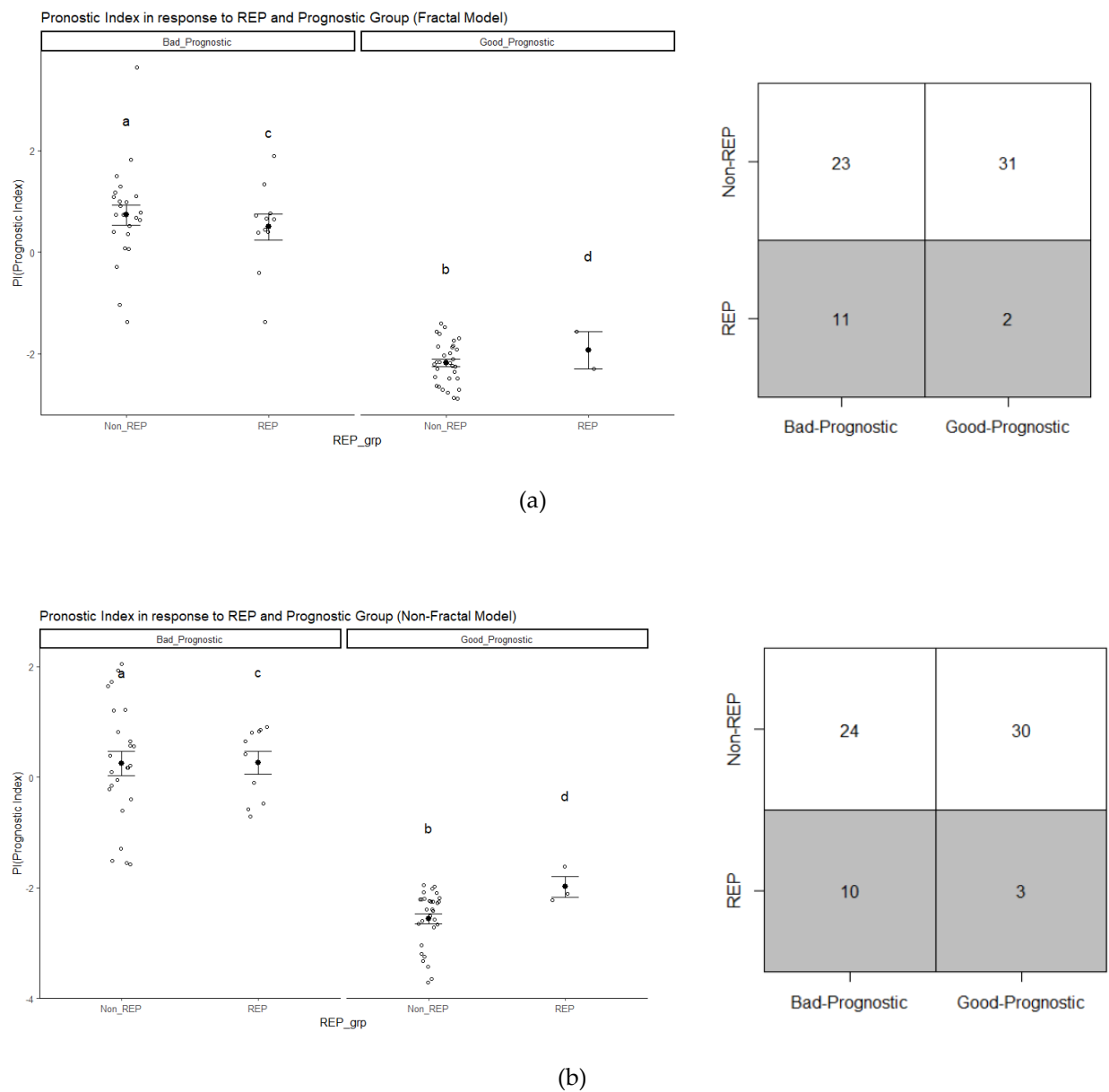

(a)

(b)

**Figure 10**. The distribution of REP progression cases in (a) Fractal, (b) non-fractal model based on prognostic group (Good or bad).

The percentage of REP cases in the bad prognostic group is higher in the fractal model compared to the non-fractal model. Therefore, we calculate whether there is significant difference between groups in survival time as depicted in Table 12. In both groups, there is a significant (p-value<0.05) difference in survival time. Test of normality and appropriate test (ANOVA/ Wilcoxon-Mann-Whitney) are performed between the two groups. From Table 12 we observe the median survival days for REP is 172 days compared to non-REP group 474.50 days and there is significant difference between the survival days between them. Moreover, in each prognostic group (good or bad) we also observe a significant difference between the groups in survival days. The median survival days for bad prognostic group is 329 days which is significantly different from the median survival days of 511 days in good prognostic group.

In addition, we analyze each individual significant feature of the fractal model with respect to REP status. For each individual feature, a test of normality, Shapiro-Wilk, is performed, and afterwards an appropriate test (ANOVA/ Wilcoxon-Mann-Whitney) is performed to determine the significance of each feature with REP status. From the Figure 11, we observe that 2 features from fractal survival probability are significantly (p-value<0.05) associated with REP status. Among the 2 features, the fractal features belong to the top 8 features of fractal model. Therefore, the higher percentage of REP cases in fractal model can be associated with the significance of this feature with REP status. The multifractal feature from T1C is significantly associated with the dependent censoring survival probability and REP status.

**Table 12.** Statistical analysis of Survival time (includes censoring time) in Prognostic and REP groups for fractal model.

| Group Name | Number of Cases | Mean | Standard Deviation | Standard Error | Median | Range | p-value |
|---|---|---|---|---|---|---|---|
| Bad Prognostic | 34 | 420.382 | 335.353 | 57.513 | 329.00 | 48.00 - 1211.00 | 0.02 |
| Good Prognostic | 33 | 678.424 | 494.209 | 86.031 | 511.00 | 57.00 – 1821.00 | |
| Non-REP | 54 | 604.259 | 441.961 | 60.143 | 474.50 | 48.00 – 1821.00 | 0.006 |
| REP | 13 | 311.615 | 340.626 | 94.472 | 172.00 | 57.00 – 1211.00 | |

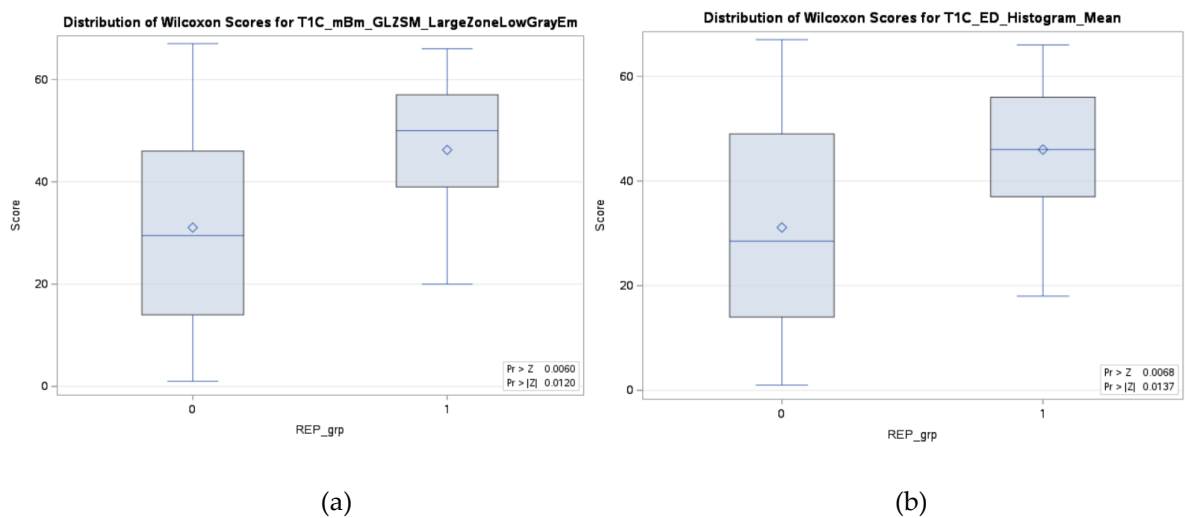

(a) (b)

**Figure 11.** Two significant (p-value<0.05) features for both survival probability under dependent censoring and REP progression status in fractal model.

## 4. Discussions

This study proposes feasibilty of radiomics and sophisticated multi-resolutional fractal texture features for prediction of REP status in GB patients from radiation-planning T1C sequence MRI. Two models (non-fractal and fractal) are constructed utilizing random sampling 5-fold cross validation as presented in Table 5. The predictive performance of fractal model is AUC 0.793 ± 0.082 with FPR of 0.145± 0.107 while that of non-fractal model AUC of 0.673±0.082 with FPR of 0.262±0.177. There is significant difference (p-value<0.001) between the fractal and the non-fractal model for prediction of REP status.

Further, a copula based modeling for survival analysis of dependent censoring is obtained using survival time, censoring time, and radiomics features. For binary prediction of patient survival, selected significant features (p-value<0.05) from survival analysis are incorporated. The predictive precision performance of patient survival for fractal model is 0.881±0.056 with FPR of 0.311±0.109 while that of non-fractal model is 0.872±0.054 with FPR of 0.339±0.106. Moreover, inclusion of IDH mutation status in addition to radiomics features is significant (p-value=0.04) in survival probability analysis of patients.

Note a direct comparison of our proposed method with the existing literature may not be possible because of the difference in patient dataset and the analysis methods. The methods in the literature are mostly focused on describing REP and assessing its impact on disease outcomes using pre-radiation MRI. In addition, few studies [4,45] focused on determining the significance of radiation-planning MRI for pseudo-progression assessment. Previous studies [7] investigated the intergration of REP and MGMT status for overall survival of patients. Patients with both REP and unmethylated MGMT status have worse prognosis compared to non-REP and methylated patients (10.2 months versus 16.5 months, p-value=0.033).

In comparison to the existing literature on REP in pre-radiation MRI, our method is focused on evaluating the diagnostic and prognostic ability of radiomics features in both identifying REP in patients with glioblastoma, and predicting their survival outcomes under dependent censoring. We first extracted radiomics features from radiation-planning T1C MRI and assessed their ability to predict REP. Our analysis show that inclusion of multi-resolution fractal features improves model performace significantly (p-value<0.05) when compared to non-fractal feature model. We then carried out survival anlysis utlizing copula based modeling of the occurrence of dependent censoring. The copula graphic (CG) estimator is used to straify good and bad prognostic groups based on radiomics and multi-resolution fractal feature-based PI. Median survival in days was higher for good versus poor prognostic group (511.0 versus 329.0; p-value =0.02). Similarily, median survival was higher in non-REP versus REP group (474.5 versus 172.0; p-value 0.006). Moreover, a fractal texture feature (T1C_mBm_GLZSM_LargeZoneLowGrayEmphasis) is found significant (p-value<0.05) along with histogram mean (edema tumor region) feature for stratification of prognostic grouping as presented in Figure 11. As far as incorporating molecular data, the inclusion of IDH mutation status with radiomics features was significantly associated with survival as shown in Figure 9. Despite being associated with outcomes in prior report[15,46,47] our analysis show that MGMT promoter methylation status is not associated with survival outcomes (p-value=0.9651) similar to that in Ref[8]. This may be explained by the fact that MGMT promoter methylation was missing in 14 patients.

Our findings are consistent with earlier research employing sophisticated imaging features, including conventional shape, volumetric, histogram statistics, texture, fractal based texture features, radiogenomics in brain tumor segmentation, classification, survival prediction and molecular mutation characterization [10,29,48–52]. Several studies focus on the application of machine or deep learning model for survival predicition, and psudeoprogression prediction with radimics features extracted from structural or advanced MRI [10,45,53–55].The inclusion of fractal based features with conventional radiomics features in the fractal model increases predictive performance significantly compared to only conventional radiomics features in the non-fractal model. The feasibility of using fractal based features is also observed in prognostic grouping using copula graphic estimator. The percentage of REP cases is higher (84.62%, 11 out of 13) with fractal features based PI when compared to non-fractal feature (76.92%, 10 out of 13) based PI. Futhermore, multi-fractal based texture feature extracted from T1C is signifiant in prognostic grouping and REP status stratification.

Limitations of our study include the indeterminate status of molecular markers for some ptients. The molecular marker of 1p/19q co-deletion status is diagnostic for oligodendroglioma which is a different type of glioma, so it is not included in our analysis. The indeterminate status of MGMT methylation and IDH mutation may also impact the analysis of these molecular marker. Moreover, the sample size of our study is rather small and may restrict the generalizability of the study. Attempts are made to address these challenges. Random sampling with 5-fold cross-validations is performed for feature selection and model evaluation to balance number of patient cases in each iterations.. In addition, only statistically significant features are included in REP classification and survival modeling. For survival analysis, copula modeling is utilized to circumvent independent censoring assumption.

Future studies with larger sample size and MRIs from different institutions/scanners are needed for improved generalizability of our model. Including additional sequences (i.e, T2/FLAIR) in addition to T1C may improve REP modeling, both in pre-operative and radiation planning scans. The ability to predict who will develop rapid progression and where spatially using pre-operative scans would be helpful in informing surgical and radiation planning decisions.

## 5. Conclusions

The most aggressive glioma in adults is glioblastoma and rapid-early progression is a negative prognostic factor. Our study demonstrates that utilizing conventional and sophisticated multi-resolution fractal image features from radiation-planning T1C MRI sequence provide a useful tool for predicting REP in glioblastoma patients. Additionally, copula based feature selection modeling and with survival analysis under dependent censoring indicate the feasibility of fractal and radiomis features to predict REP in GB patients utilizing radiation-planning MRIs.

**Author Contributions:** Conceptualization: K.I, J.P, M.H, M.M.B, and W.F.; methodology: K.I, W.F, N.D; software: W.F, N.D. ; validation: K.I, N.D and Z.A.S,M.M.B,M.H, J.P ; formal analysis: W.F, N.D, K.I ; investigation: W.F, K.I, N.D.; data curation: M.H, M.M.B, M.M.L,S.D,K.I and W.F.; writing—original draft preparation: W.F; writing—review and editing: K.I, N.D. Z.A.S, M.H, M.M.B, M.M.L, J.P,S.D. ; visualization: W.F ; supervision: K.I, N.D., M.M.B, J.P,M.H; project administration: K. I., M.M.L; All authors have read and agreed to the published version of the manuscript.

**Funding:** The research received no external funding.

**Institutional Review Board Statement:** The study was conducted in accordance with the Declaration of Helsinki and approved by the Institutional Review Board of Old Dominion University.

**Informed Consent Statement:** Informed consent was obtained from all subjects involved in the study.

**Data Availability Statement:** Data may be available subject to appropriate IRB approval.

**Acknowledgments:** The authors thank Jay Anderson, Amanda Magee and Ben Purugganan for assistance in patients' data collection and administrative support.

**Conflicts of Interest:** The authors declared no conflict of interest.

**References**

1. Miller, K.D.; Ostrom, Q.T.; Kruchko, C.; Patil, N.; Tihan, T.; Cioffi, G.; Fuchs, H.E.; Waite, K.A.; Jemal, A.; Siegel, R.L.; et al. Brain and Other Central Nervous System Tumor Statistics, 2021. *CA. Cancer J. Clin.* **2021**, *71*, 381–406, doi:10.3322/caac.21693.

2. Nam, J.Y.; de Groot, J.F. Treatment of Glioblastoma. *J. Oncol. Pract.* **2017**, *13*, 629–638, doi:10.1200/JOP.2017.025536.

3. Stupp, R.; Taillibert, S.; Kanner, A.; Read, W.; Steinberg, D.M.; Lhermitte, B.; Toms, S.; Idbaih, A.; Ahluwalia, M.S.; Fink, K.; et al. Effect of Tumor-Treating Fields Plus Maintenance Temozolomide vs Maintenance Temozolomide Alone on Survival in Patients With Glioblastoma: A Randomized Clinical Trial. *JAMA* **2017**, *318*, 2306–2316, doi:10.1001/jama.2017.18718.

4. Majós, C.; Cos, M.; Castañer, S.; Pons, A.; Gil, M.; Fernández-Coello, A.; Maciá, M.; Bruna, J.; Aguilera, C. Preradiotherapy MR Imaging: A Prospective Pilot Study of the Usefulness of Performing an MR Examination Shortly before Radiation Therapy in Patients with Glioblastoma. *Am. J. Neuroradiol.* **2016**, *37*, doi:10.3174/ajnr.A4917.

5. Pirzkall, A.; McGue, C.; Saraswathy, S.; Cha, S.; Liu, R.; Vandenberg, S.; Lamborn, K.R.; Berger, M.S.; Chang, S.M.; Nelson, S.J. Tumor Regrowth between Surgery and Initiation of Adjuvant Therapy in Patients with Newly Diagnosed Glioblastoma. *Neuro. Oncol.* **2009**, *11*, doi:10.1215/15228517-2009-005.

6. Farace, P.; Amelio, D.; Ricciardi, G.K.; Zoccatelli, G.; Magon, S.; Pizzini, F.; Alessandrini, F.; Sbarbati, A.; Amichetti, M.; Beltramello, A. Early MRI Changes in Glioblastoma in the Period between Surgery and Adjuvant Therapy. *J. Neurooncol.* **2013**, *111*, doi:10.1007/s11060-012-0997-y.

7. Palmer, J.D.; Bhamidipati, D.; Shukla, G.; Sharma, D.; Glass, J.; Kim, L.; Evans, J.J.; Judy, K.; Farrell, C.; Andrews, D.W.; et al. Rapid Early Tumor Progression Is Prognostic in Glioblastoma Patients. *Am. J. Clin. Oncol. Cancer Clin. Trials* **2019**, *42*, 481–486, doi:10.1097/COC.0000000000000537.

8. Lakomy, R.; Kazda, T.; Selingerova, I.; Poprach, A.; Pospisil, P.; Belanova, R.; Fadrus, P.; Smrcka, M.; Vybihal, V.; Jancalek, R.; et al. Pre-Radiotherapy Progression after Surgery of Newly Diagnosed Glioblastoma: Corroboration of New Prognostic Variable. *Diagnostics* **2020**, *10*, 1–13, doi:10.3390/diagnostics10090676.

9. NCCN Guidelines. NCCN Guidelines for Treatment of Cancer by Site. Central Nervous System Cancers. National Comprehensive Cancer Network. Available online: https://www.nccn.org/guidelines/category_1#cns (accessed on 21 February 2023).

10. Kazerooni, A.F.; Bagley, S.J.; Akbari, H.; Saxena, S.; Bagheri, S.; Guo, J.; Chawla, S.; Nabavizadeh, A.; Mohan, S.; Bakas, S.; et al. Applications of Radiomics and Radiogenomics in High-Grade Gliomas in the Era of Precision Medicine. *Cancers (Basel).* **2021**, *13*, 1–15, doi:10.3390/cancers13235921.

11. Mulford, K.; McMahon, M.; Gardeck, A.M.; Hunt, M.A.; Chen, C.C.; Odde, D.J.; Wilke, C. Predicting Glioblastoma Cellular Motility from In Vivo MRI with a Radiomics Based Regression Model. *Cancers (Basel).* **2022**, *14*, doi:10.3390/cancers14030578.

12. Cepeda, S.; Pérez-Nuñez, A.; García-García, S.; García-Pérez, D.; Arrese, I.; Jiménez-Roldán, L.; García-Galindo, M.; González, P.; Velasco-Casares, M.; Zamora, T.; et al. Predicting Short-Term Survival after Gross Total or near Total Resection in Glioblastomas by Machine Learning-Based Radiomic Analysis of Preoperative Mri. *Cancers (Basel).* **2021**, *13*, doi:10.3390/cancers13205047.

13. Pereira, S.; Pinto, A.; Alves, V.; Silva, C.A. Brain Tumor Segmentation Using Convolutional Neural Networks in MRI Images. *IEEE Trans. Med. Imaging* **2016**, *35*, 1240–1251, doi:10.1109/TMI.2016.2538465.

14. Pei, L.; Vidyaratne, L.; Rahman, M.M.; Iftekharuddin, K.M. Context Aware Deep Learning for Brain Tumor Segmentation, Subtype Classification, and Survival Prediction Using Radiology Images. *Sci. Rep.* **2020**, *10*, 19726, doi:10.1038/s41598-020-74419-9.

15. Shboul, Z.A.; Chen, J.; M. Iftekharuddin, K. Prediction of Molecular Mutations in Diffuse Low-Grade Gliomas Using MR Imaging Features. *Sci. Rep.* **2020**, *10*, 1–13, doi:10.1038/s41598-020-60550-0.

16. Emura, T.; Chen, Y.-H. *Analysis of Survival Data with Dependent Censoring Copula-Based Approaches*; Springer Singapore: Singapore, 2018; ISBN 9789811071638.

17. Emura, T.; Chen, Y.H. Gene Selection for Survival Data under Dependent Censoring: A Copula-Based Approach. *Stat. Methods Med. Res.* **2016**, *25*, 2840–2857, doi:10.1177/0962280214533378.

18. Waqar, M.; Trifiletti, D.M.; McBain, C.; O'Connor, J.; Coope, D.J.; Akkari, L.; Quinones-Hinojosa, A.; Borst, G.R. Early Therapeutic Interventions for Newly Diagnosed Glioblastoma: Rationale and Review of the Literature. *Curr. Oncol. Rep.* **2022**, *24*, 311–324, doi:10.1007/s11912-021-01157-0.

19. Waqar, M.; Roncaroli, F.; Lehrer, E.J.; Palmer, J.D.; Villanueva-meyer, J.; Braunstein, S.; Hall, E.; Aznar, M.; Hamer, P.C.D.W.; Urso, P.I.D.; et al. Neuro-Oncology Advances Systematic Review and Meta-Analysis. **2022**, *4*, 1–10.

20. Smith, S.M.; Jenkinson, M.; Woolrich, M.W.; Beckmann, C.F.; Behrens, T.E.J.; Johansen-Berg, H.; Bannister, P.R.; De Luca, M.; Drobnjak, I.; Flitney, D.E.; et al. Advances in Functional and Structural MR Image Analysis and Implementation as FSL. *Neuroimage* **2004**, *23 Suppl 1*, S208-19, doi:10.1016/j.neuroimage.2004.07.051.

21. Smith, S.M. Fast Robust Automated Brain Extraction. *Hum. Brain Mapp.* **2002**, *17*, 143–155, doi:10.1002/hbm.10062.

22. Smith, S.M.; Brady, J.M. SUSAN - A New Approach to Low Level Image Processing. *Int. J. Comput. Vis.* **1997**, *23*, 45–78, doi:10.1023/A:1007963824710.

23. Sled, J.G.; Zijdenbos, A.P.; Evans, A.C. A Nonparametric Method for Automatic Correction of Intensity Nonuniformity in MRI Data. *IEEE Trans. Med. Imaging* **1998**, *17*, 87–97, doi:10.1109/42.668698.

24. Menze, B.H.; Jakab, A.; Bauer, S.; Kalpathy-Cramer, J.; Farahani, K.; Kirby, J.; Burren, Y.; Porz, N.; Slotboom, J.; Wiest, R.; et al. The Multimodal Brain Tumor Image Segmentation Benchmark (BRATS). *IEEE Trans. Med. Imaging* **2015**, *34*, 1993–2024, doi:10.1109/TMI.2014.2377694.

25. Baid, U.; Ghodasara, S.; Bilello, M.; Mohan, S.; Calabrese, E.; Colak, E.; Farahani, K.; Kalpathy-Cramer, J.; Kitamura, F.C.; Pati, S.; et al. The RSNA-ASNR-MICCAI BraTS 2021 Benchmark on Brain Tumor Segmentation and Radiogenomic Classification. **2021**.

26. Bakas, S.; Akbari, H.; Sotiras, A.; Bilello, M.; Rozycki, M.; Kirby, J.S.; Freymann, J.B.; Farahani, K.; Davatzikos, C. Advancing The Cancer Genome Atlas Glioma MRI Collections with Expert Segmentation Labels and Radiomic Features. *Sci. Data* **2017**, *4*, 2017–2018, doi:10.1038/sdata.2017.117.

27. Pei, L.; Vidyaratne, L.; Rahman, M.M.; Iftekharuddin, K.M. Context Aware Deep Learning for Brain Tumor Segmentation, Subtype Classification, and Survival Prediction Using Radiology Images. *Sci. Rep.* **2020**, *10*, 1–12, doi:10.1038/s41598-020-74419-9.

28. Iftekharuddin, K.M.; Jia, W.; Marsh, R. Fractal Analysis of Tumor in Brain MR Images. *Mach. Vis. Appl.* **2003**, *13*, 352–362, doi:10.1007/s00138-002-0087-9.

29. Reza, S.M.S.; Mays, R.; Iftekharuddin, K.M. Multi-Fractal Detrended Texture Feature for Brain Tumor Classification. In Proceedings of the Medical Imaging 2015: Computer-Aided Diagnosis; 2015; Vol. 9414.

30. Ayache, A.; Véhel, J.L. On the Identification of the Pointwise Hölder Exponent of the Generalized Multifractional

Brownian Motion. *Stoch. Process. their Appl.* **2004**, *111*, doi:10.1016/j.spa.2003.11.002.

31. John, P. Brain Tumor Classification Using Wavelet and Texture Based Neural Network. *Int. J. Sci. Eng. Res.* **2012**, *3*.

32. Farzana, W.; Shboul, Z.A.; Temtam, A.; Iftekharuddin, K.M. Uncertainty Estimation in Classification of MGMT Using Radiogenomics for Glioblastoma Patients. In Proceedings of the Proc. SPIE 12033, Medical Imaging 2022: Computer-Aided Diagnosis; 2022; p. 51.

33. Kamalov, F.; Thabtah, F.; Leung, H.H. Feature Selection in Imbalanced Data. *Ann. Data Sci.* **2022**, doi:10.1007/s40745-021-00366-5.

34. Cox, D.R. Regression Models and Life-Tables. *J. R. Stat. Soc. Ser. B* **1972**, *34*, 187–220.

35. Emura, T.; Chen, Y.H.; Chen, H.Y. Survival Prediction Based on Compound Covariate under Cox Proportional Hazard Models. *PLoS One* **2012**, *7*, doi:10.1371/journal.pone.0047627.

36. Emura, T.; Matsui, S.; Chen, H.Y. Compound.Cox: Univariate Feature Selection and Compound Covariate for Predicting Survival. *Comput. Methods Programs Biomed.* **2019**, *168*, 21–37, doi:10.1016/j.cmpb.2018.10.020.

37. Statistics in Medicine - 2018 - Weber - Quantifying the Association between Progression-free Survival and Overall Survival.Pdf.

38. Harrell, F.E.; Califf, R.M.; Pryor, D.B.; Lee, K.L.; Rosati, R.A. Evaluating the Yield of Medical Tests. *JAMA J. Am. Med. Assoc.* **1982**, *247*, doi:10.1001/jama.1982.03320430047030.

39. Harrell, F.E.; Lee, K.L.; Mark, D.B. Multivariable Prognostic Models: Issues in Developing Models, Evaluating Assumptions and Adequacy, and Measuring and Reducing Errors. *Stat. Med.* **1996**, *15 4*, 361–387.

40. Schweizer, B. Introduction to Copulas. *J. Hydrol. Eng.* **2007**, *12*, doi:10.1061/(asce)1084-0699(2007)12:4(346).

41. Chen, Y.-H. Semiparametric Marginal Regression Analysis for Dependent Competing Risks under an Assumed Copula. *J. R. Stat. Soc. Ser. B (Statistical Methodol.* **2010**, *72*, 235–251.

42. Rivest, L.-P.; Wells, M.T. A Martingale Approach to the Copula-Graphic Estimator for the Survival Function under Dependent Censoring. *J. Multivar. Anal.* **2001**, *79*, 138–155, doi:https://doi.org/10.1006/jmva.2000.1959.

43. Pepe, M.S.; Fleming, T.R. Weighted Kaplan-Meier Statistics: A Class of Distance Tests for Censored Survival Data. *Biometrics* **1989**, *45*, 497–507.

44. Frankel, P.H.; Reid, M.E.; Marshall, J.R. A Permutation Test for a Weighted Kaplan-Meier Estimator with Application to the Nutritional Prevention of Cancer Trial. *Contemp. Clin. Trials* **2007**, *28*, 343–347, doi:10.1016/j.cct.2006.10.006.

45. Baine, M.; Burr, J.; Du, Q.; Zhang, C.; Liang, X.; Krajewski, L.; Zima, L.; Rux, G.; Zhang, C.; Zheng, D. The Potential Use of Radiomics with Pre-Radiation Therapy Mr Imaging in Predicting Risk of Pseudoprogression in Glioblastoma Patients. *J. Imaging* **2021**, *7*, doi:10.3390/jimaging7020017.

46. Houdova Megova, M.; Drábek, J.; Dwight, Z.; Trojanec, R.; Koudeláková, V.; Vrbková, J.; Kalita, O.; Mlcochova, S.; Rabcanova, M.; Hajdúch, M. Isocitrate Dehydrogenase Mutations Are Better Prognostic Marker than O6-Methylguanine-DNA Methyltransferase Promoter Methylation in Glioblastomas - a Retrospective, Single-Centre Molecular Genetics Study of Gliomas. *Klin. Onkol.* **2017**, *30*, 361–371, doi:10.14735/amko2017361.

47. Killela, P.J.; Pirozzi, C.J.; Healy, P.; Reitman, Z.J.; Lipp, E.; Rasheed, B.A.; Yang, R.; Diplas, B.H.; Wang, Z.; Greer, P.K.; et al. Mutations in IDH1, IDH2, and in the TERT Promoter Define Clinically Distinct Subgroups of Adult Malignant Gliomas. *Oncotarget* **2014**, *5*, 1515–1525, doi:10.18632/oncotarget.1765.

48. Shboul, Z.A.; Alam, M.; Vidyaratne, L.; Pei, L.; Elbakary, M.I.; Iftekharuddin, K.M. Feature-Guided Deep Radiomics for Glioblastoma Patient Survival Prediction. *Front. Neurosci.* **2019**, *13*, 1–17, doi:10.3389/fnins.2019.00966.

49. Shboul, Z.; Vidyaratne, L.; Alam, M.; Reza, S.M.S.; Iftekharuddin, K.M. Glioblastoma and Survival Prediction. *Brainlesion glioma, Mult. sclerosis, stroke Trauma. brain Inj. BrainLes* **2018**, *10670*, 358–368, doi:10.1007/978-3-319-75238-9_31.

50. Bakas, S.; Reyes, M.; Jakab, A.; Bauer, S.; Rempfler, M.; Crimi, A.; Shinohara, R.T.; Berger, C.; Ha, S.M.; Rozycki, M.; et al. Identifying the Best Machine Learning Algorithms for Brain Tumor Segmentation, Progression Assessment, and Overall Survival Prediction in the BRATS Challenge. **2018**.

51. Shboul, Z.A.; Diawara, N.; Vossough, A.; Chen, J.Y.; Iftekharuddin, K.M. Joint Modeling of RNAseq and Radiomics Data for Glioma Molecular Characterization and Prediction. *Front. Med.* **2021**, *8*, doi:10.3389/fmed.2021.705071.

52. Pei, L.; Vidyaratne, L.; Monibor Rahman, M.; Shboul, Z.A.; Iftekharuddin, K.M. Multimodal Brain Tumor Segmentation and Survival Prediction Using Hybrid Machine Learning. In Proceedings of the Brainlesion: Glioma, Multiple Sclerosis, Stroke and Traumatic Brain Injuries; Crimi, A., Bakas, S., Eds.; Springer International Publishing: Cham, 2020; pp. 73–81.

53. Lohmann, P.; Elahmadawy, M.A.; Gutsche, R.; Werner, J.M.; Bauer, E.K.; Ceccon, G.; Kocher, M.; Lerche, C.W.; Rapp, M.; Fink, G.R.; et al. Fet Pet Radiomics for Differentiating Pseudoprogression from Early Tumor Progression in Glioma Patients Post-chemoradiation. *Cancers (Basel).* **2020**, *12*, 1–17, doi:10.3390/cancers12123835.

54. Chiu, F.Y.; Yen, Y. Efficient Radiomics-Based Classification of Multi-Parametric MR Images to Identify Volumetric Habitats and Signatures in Glioblastoma: A Machine Learning Approach. *Cancers (Basel).* **2022**, *14*, doi:10.3390/cancers14061475.

55. Yoon, H.G.; Cheon, W.; Jeong, S.W.; Kim, H.S.; Kim, K.; Nam, H.; Han, Y.; Lim, D.H. Multi-Parametric Deep Learning Model for Prediction of Overall Survival after Postoperative Concurrent Chemoradiotherapy in Glioblastoma Patients. *Cancers (Basel).* **2020**, *12*, 1–12, doi:10.3390/cancers12082284.